\documentclass[sigconf]{acmart}
\AtBeginDocument{%
	\providecommand\BibTeX{{%
			\normalfont B\kern-0.5em{\scshape i\kern-0.25em b}\kern-0.8em\TeX}}}

\usepackage{filecontents}
\usepackage{latexsym}
\usepackage{balance}
\usepackage{flushend}

\usepackage{microtype}
\usepackage{algorithm,algpseudocode}
\usepackage{graphicx}
\usepackage{bm}
\usepackage{amsmath}
\usepackage{subfigure}
\usepackage{makecell}
\usepackage{multirow}
\usepackage{cleveref}
\usepackage{hyperref}
\hypersetup{
	colorlinks,
	citecolor=gray,
	linkcolor=red,
	urlcolor=blue}

\copyrightyear{2023}
\acmYear{2023}
\setcopyright{acmlicensed}\acmConference[KDD '23]{Proceedings of the 29th
	ACM SIGKDD Conference on Knowledge Discovery and Data Mining}{August
	6--10, 2023}{Long Beach, CA, USA}
\acmBooktitle{Proceedings of the 29th ACM SIGKDD Conference on Knowledge
	Discovery and Data Mining (KDD '23), August 6--10, 2023, Long Beach, CA,
	USA}
\acmPrice{15.00}
\acmDOI{10.1145/3580305.3599284}
\acmISBN{979-8-4007-0103-0/23/08}

\begin{document}
	\newcommand{\sysname}{{ToP}}
	\newcommand{\lz}[1]{{\textcolor{red}{\it Lyna: #1}}}	
		\title{Constraint-aware and Ranking-distilled  Token Pruning for Efficient Transformer Inference}
\author{Junyan Li}
\authornote{Work was done during the internship at Microsoft Research}
\affiliation{%
	\institution{Zhejiang University}
	\country{lijunyan668@outlook.com\vspace{-0.3em}}
}
\author{Li Lyna Zhang}
\authornote{Corresponding author}
\affiliation{%
	\institution{Microsoft Research}
	\country{lzhani@microsoft.com\vspace{-0.3em}}
}
\author{Jiahang Xu}
\affiliation{%
	\institution{Microsoft Research}
	\country{jiahangxu@microsoft.com\vspace{-0.3em}}
}
\author{Yujing Wang}
\affiliation{%
	\institution{Microsoft}
	\country{yujwang@microsoft.com\vspace{-0.3em}}
}
\author{Shaoguang Yan}
\affiliation{%
	\institution{Microsoft}
	\country{shaoyan@microsoft.com\vspace{-0.3em}}
}
\author{Yunqing Xia}
\affiliation{%
	\institution{Microsoft}
	\country{yxia@microsoft.com\vspace{-0.3em}}
}
\author{Yuqing Yang}
\affiliation{%
	\institution{Microsoft Research}
	\country{yuqing.yang@microsoft.com\vspace{-0.3em}}
}
\author{Ting Cao}
\affiliation{%
	\institution{Microsoft Research}
	\country{ting.cao@microsoft.com\vspace{-0.3em}}
}
\author{Hao Sun}
\affiliation{%
	\institution{Microsoft}
	\country{hasun@microsoft.com\vspace{-0.3em}}
}
\author{Weiwei Deng}
\affiliation{%
	\institution{Microsoft}
	\country{dedeng@microsoft.com}
}
\author{Qi Zhang}
\affiliation{%
	\institution{Microsoft}
	\country{zhang.qi@microsoft.com}
}
\author{Mao Yang}
\affiliation{%
	\institution{Microsoft Research}
	\country{maoyang@microsoft.com}
}

\renewcommand{\shortauthors}{Junyan Li  et al.}
	\begin{abstract}	
	Deploying pre-trained transformer models like BERT on downstream tasks in resource-constrained scenarios is challenging due to their high inference cost, which grows rapidly with input sequence length.  	In this work, we propose a constraint-aware and ranking-distilled token pruning method \textbf{\sysname}, which  selectively removes unnecessary tokens as input sequence passes through layers, allowing the model to improve online inference speed while preserving accuracy.  {{\sysname}} overcomes the limitation of inaccurate token importance ranking in the conventional self-attention mechanism through a ranking-distilled token distillation technique, which distills effective token rankings from the final layer of unpruned models to early layers of pruned models.
	Then, {\sysname} introduces a  coarse-to-fine pruning approach that automatically selects the optimal subset of transformer layers and optimizes token pruning decisions within these layers through improved $L_0$ regularization. Extensive experiments on GLUE benchmark and SQuAD tasks demonstrate that {\sysname} outperforms state-of-the-art token pruning and model compression methods with improved accuracy and speedups. 	{\sysname} reduces the average FLOPs of BERT by 8.1$\times$ while achieving competitive accuracy on GLUE, and provides a real latency speedup of up to 7.4$\times$ on an Intel CPU. Code is available at \href{https://github.com/microsoft/Moonlit/tree/main/ToP}{\textcolor{purple}{here}~\footnote{\url{https://github.com/microsoft/Moonlit/tree/main/ToP}}}. 
\vspace{-0.5ex}
	

\end{abstract}
\maketitle
\section{Introduction}
Pre-trained transformer models~\cite{bert,albert,roberta,t5} have achieved great success for a wide variety of NLP tasks. 
However, the superior performance comes at the cost of increasingly larger model sizes and computation overhead, making it difficult to efficiently deploy them on different downstream tasks in various latency-critical scenarios such as online servers and edge devices.  


Accelerating transformer inference is often achieved through model compression methods such as pruning~\cite{movement,nn_pruning}, quantization~\cite{shen2020q,kim2021bert,chen2021quantization}, and knowledge distillation~\cite{sanh2020distilbert,jiao2020tinybert}. These techniques aim to reduce the size of the model, with quantization and distillation resulting in a smaller, fixed model. Structured pruning, which eliminates redundant heads or dimensions, can effectively meet deployment requirements~\cite{swiftpruner,cofi}. However, structured pruning may not guarantee optimal accuracy, particularly for small transformers or long input sequences, as the attention mechanism has a $O(n^2)$ computation complexity with input token length $n$. This means a significant portion of the model must be pruned to meet tight deployment constraints, potentially compromising accuracy.

 

Recently, a promising subfield in NLP has emerged that focuses on reducing latency during model inference by pruning input tokens.  It's based on the intuition that not all tokens in the input sequence are critical for making a final prediction. As  tokens pass through the encoder layers,  some tokens have been captured by other tokens via attention in the early layer and do not require future modeling in a higher layer~\cite{rogers-etal-2020-primer,powerbert}. Pruning these uninformative tokens within each layer can increase the model's inference speed without sacrificing accuracy. Moreover, the removal of these tokens in each layer will also reduce the computation and memory requirements in its subsequent layers, resulting in linear or even quadratic reductions and providing greater compression benefits.


Some prior works~\cite{powerbert,LAT,ltp,transkimmer,trbert} have examined the potential of layer-wise token pruning of input sequences.  However, these approaches face several limitations. First,  they treat all layers equally, leading to a vast design space, as pruning decisions must be made for each token at every layer through the use of token-level masks.
 Second, existing methods primarily aim to minimize accuracy drop and  use regularization loss terms to encourage maximum token pruning~\cite{powerbert,ltp,transkimmer}, lacking effective control over the given token sparsity ratio. This can be problematic in real-world scenarios where a specific sparsity ratio or deployment constraint is often required.
 

 
 Finally, existing token importance scoring criterion struggles to achieve both high accuracy and real inference efficiency. Attention value-based approaches~\cite{powerbert,ltp}, which utilize the self-attention mechanism to score token importance,  can be efficient, but may inadvertently remove important tokens that receive little attention in early layers~\cite{rogers-etal-2020-primer}, leading to a huge drop in accuracy.  On the other hand, prediction module-based approaches~\cite{transkimmer,trbert}, which insert extra neural networks to predict token importance scores, can be more accurate. But they also come with the cost of introducing considerable additional inference cost (i.e., $\sim30\%$ latency overhead on GPU) and may impede overall speedup. Given the tight latency constraints in  many real-world deployments, attention value-based scoring is a more promising method for achieving efficiency. However, the challenge of improving accuracy for determining token importance in early layers remains unresolved.
 
 In our work, we introduce \textbf{To}\textbf{P} (\textbf{To}ken \textbf{P}runing), a deployment-friendly and constraint-aware token pruning approach that addresses all the above challenges. {\sysname} trains an optimal token pruning decision based on our improved attention value-based scoring, enabling dynamic removal of unnecessary tokens layer-by-layer during inference, while preserving accuracy and meeting deployment constraints. Our approach incorporates two key techniques.
 
 First, we introduce a new token distillation method called \textit{ranking-aware token distillation} to enhance the ability of self-attention values to rank token importance in early layers, thereby resolving the issue of unintended removal of top-ranked tokens  during inference which can affect the model's accuracy. Our solution is inspired by the observation that attention values in deeper layers rank token importance more accurately. We thus utilize the importance rankings generated by the final layers of the unpruned model as knowledge and distill it to the early layers.
 Conventional distillation methods~\cite{cofi,Sun2019PatientKD} commonly use the MSE loss to measure the layer-wise representations' difference  between teacher and student, but this may not be effective for transformer models such as BERT, which capture different levels of information across early and deep layers. Directly minimizing the MSE loss of absolute attention values may lead to suboptimal results. Instead, our method proposes a ranking-aware distillation loss that minimizes the differences in token importance \textit{rankings} between the final layer of the teacher and the early layers of the student using the LambdaRank loss~\cite{lambdandcg}. This distillation effectively retains the most important tokens and results in significant accuracy improvements.

Next, we present a generic learning algorithm that optimizes token pruning decisions based on our improved attention value-based importance scoring. Different from prior works, we utilize two-tier binary masks, consisting of coarse-grained gate masks and fine-grained token ranking masks, to automatically determine the optimal subset of transformer layers for fine-grained token pruning. The gate masks act as layer selectors, while the ranking masks dynamically identify which sorted tokens (\textit{i.e.,} based on the attention value scores) within the selected layers to be pruned. This design allows for a more flexible pruning space and eases the optimization compared to learning all token masks equally. 
To find the optimal mask values while achieving the desired pruning ratio, we solve an end-to-end optimization problem using improved $L_0$ regularization~\cite{lagrangian}, which  jointly learns these masks and updates model parameters on the target downstream tasks, resulting in better model accuracy.  
We summarize our contributions as follows:
\begin{itemize}
	\item For the first time, we propose ranking-aware token distillation to  effectively improve token importance rankings based on attention values, which  greatly enhances the effectiveness of token pruning methods relying on attention values.
	\item We further propose a constraint-aware token pruning algorithm ({\sysname}). For a given deployment constraint, {\sysname} automatically selects the optimal subset of transformer layers and optimizes token pruning decisions within these layers through improved $L_0$ regularization.
	
	\item Extensive experiments on GLUE benchmark~\cite{glue} and SQuAD v2.0~\cite{squadv2} demonstrate that {\sysname} consistently outperform  state-of-the-art token pruning and model compression baselines with higher accuracy and speedups. By removing unnecessary tokens, {\sysname} improves accuracy by up to 4.3\% on GLUE and reduces FLOPs by an average of 6.7$\times$.
	Furthermore,   {\sysname} delivers substantial real latency reduction, with up to 7.4$\times$ acceleration for BERT inference on CPU. 
\end{itemize}

\section{Related Works}
\subsection{Model Compression}
To reduce the inference cost of pre-trained transformer models, a variety of compression techniques have been proposed, including weight pruning~\cite{movement,gordon2020compressing,nn_pruning}, quantization~\cite{shen2020q,kim2021bert,chen2021quantization} and distillation~\cite{sanh2020distilbert,jiao2020tinybert}. Token-level pruning has been shown to complement knowledge distillation and quantization~\cite{ltp}.  Here, we focus on pruning and distillation and briefly discuss the related work. 

\noindent\textbf{Weight pruning} is categorized into 1) unstructured and 2) structured pruning. Unstructured methods~\cite{movement, gordon2020compressing} achieve high sparsity without accuracy drop but offer minimal latency benefits due to irregular sparse patterns.
In contrast, structured pruning removes coherent weight groups, reducing latency without special hardware support.
CoFi~\cite{cofi} achieves 10$\times$ speedup with a small accuracy drop by jointly pruning layers, attention heads, FFN, and hidden units. SwiftPruner~\cite{swiftpruner} is a latency-aware pruning method that finds  optimal layer-wise pruning policies under a given latency requirement through AutoML. However, structure pruning may result in a loss of accuracy when the deployment requirements are highly constrained and the downstream task has a long input sequence. This is because the model complexity increases quadratically with token length. When the token length is long, the original model must be compressed to a high ratio, which can cause accuracy loss. 

\noindent\textbf{Knowledge distillation}~\cite{kd,Sun2019PatientKD,Turc2019WellReadSL} aims to transfer knowledge from a large teacher model to a small student model. 
 It is well known that model pruning with a distillation objective  can significantly improve accuracy~\cite{movement,nn_pruning}.
 Common distillation objectives include cross-entropy loss for output probability distributions~\cite{kd, sanh2020distilbert} and MSE loss for layer-wise representations~\cite{cofi, Sun2019PatientKD, jiao2020tinybert}. 
  However, the combination of distillation with token pruning has not been widely explored. Our aim is to transfer the knowledge of token importance rankings from the teacher's final layer to the early layers of the student model during token pruning, which poses a new challenge and requires new distillation objective functions.
 
\vspace{-1ex}
\subsection{Token Pruning}
\vspace{-0.5ex}
Existing token pruning works can be categorized into two classes based on  token removal or retention criteria. 
The first class uses attention value-based scoring~\cite{spatten,powerbert,ltp} to identify unimportant tokens.
For instance,
SpAtten~\cite{spatten} ranks tokens using importance scores and retains the top-k highest-scoring tokens.
 PoWER-BERT~\cite{powerbert} 
 learns a layer-wise token pruning ratio, and prunes all input sequences to the same length. LTP~\cite{ltp} improves PoWER-BERT by introducing a learnable layer-wise threshold, enabling adaptive pruning length. However, these approaches rely on the effectiveness of token importance scoring. As shown in Fig.~\ref{fig:attentionscore},
 crucial tokens may receive little attention in early layers, leading them to be misclassified as redundant tokens. Removing these essential tokens can result in a drastic accuracy loss.
 
The second class of token pruning methods~\cite{trbert,transkimmer}  inserts a prediction module before each transformer layer to provide a more accurate token importance score prediction. 
Transkimmer~\cite{transkimmer} is a notable example that inserts a 2-layer MLP network at each layer as the prediction module.  However, the extra prediction module can also introduce considerable inference latency overhead, which is unfriendly on resource-limited devices.

 {\sysname} addresses all the above limitations by introducing the ranking-aware token distillation technique. Since it can effectively improve the effectiveness of attention value-based scoring, we can achieve the same or better level of model accuracy as prediction-based approaches, while also being more efficient and inference-friendly.

 In addition, these works  primarily aim to minimize accuracy loss while reducing as many numbers of tokens as possible. They introduce a regularization term with hyper-parameter $\lambda$ to balance  accuracy and efficiency. However, this approach lacks effective control on pruning tokens to a desired computation budget. 
LAT~\cite{LAT} address this with an evolutionary search strategy. 
In contrast, {\sysname} solves this limitation using a different approach - a  constraint-aware token pruning algorithm, which is an optimization-based solution.

\vspace{-2ex}
\subsection{Efficient Transformers}
\vspace{-0.5ex}
Since the computation and memory cost in self-attention is quadratic in the token length, there have been a number of attempts in designing sparse attention. 
Sparse Transformer~\cite{sparsetransformer}, Longformer~\cite{longformer}, and Big Bird~\cite{bigbird} employ sparse attention to allow the model to handle long sequences. However, these methods only reduce the CUDA memory but cannot be faster than the full attention. Other efforts~\cite{xu2021bert,hanruiwang2020hat} leverage neural architecture search to design efficient transformer models with smaller depths and fewer heads. {\sysname} is orthogonal to these techniques on the input dimension reduction.

\section{Background and motivations}
\subsection{Background}
Transformer models, such as BERT~\cite{bert}, are stacked up with  multiple encoder layers. A basic transformer layer wraps a multi-head self-attention (MHA) and feed-forward (FFN) layer with residual connection and layer normalization (LN).
Given an input sequence of $n$ tokens and hidden size of $d$, 
 the hidden state of the $i^{th}$ layer, $\bm{X}_i=(x_1, x_2,...x_n)\in\mathbb{R}^{n\times d }$, is computed from the previous layer:
\begin{equation}
	\label{eq:transformer}
	\begin{aligned}
		\displaystyle \bm{X}_{i-1}^\prime=\text{LN}(\bm{X}_{i-1}+\text{MHA}(\bm{X}_{i-1}))
		\\
		\bm{X}_i=\text{LN}(\bm{X}_{i-1}^\prime+\text{FFN}(\bm{X}_{i-1}^\prime)),
	\end{aligned} 
\end{equation}

Specifically, 
an MHA consists of $N_h$ heads, where each head $h$ is associated with query $\bm{W}_q^h$, key $\bm{W}_k^h$ and value $\bm{W}_v^h$ matrix. Each head $h$ first computes an attention probability matrix $\bm{A}_h$ and then computes the self-attention mechanism as follows:
\begin{equation}
	\label{eq:attention}
	\begin{aligned}
		\displaystyle \bm{A}_h=\text{softmax}((\bm{X}\times \bm{W}_q^h )\times(\bm{X}\times\bm{W}_k^h)^T)\\
		\text{Attention}_h=\bm{A}_h\times (\bm{X}\times\bm{W}^h_v)
	\end{aligned} 
\end{equation}

\noindent\textbf{Complexity analysis}. The above self-attention measures the pairwise importance of each token on every other token in the input, and the total complexity of MHA layer is $O(d^2n+n^2d)$, which is quadratic with $n$. For FFN layer, the computation complexity is $O(nd^2)$, which is linear with $n$. When applied to long input sequences (i.e., a large $n$), the computation and memory of MHA layer grow quadratically and become very expensive. 

To address this limitation, our work introduces token pruning, where unimportant tokens are gradually dropped as the inference proceeds. For each transformer layer, which initially has $n$ tokens,  we aim to remove a specific number of unimportant tokens from them.  These removed tokens will not be considered in subsequent layers. This leads to a linear (for FFN) or quadratic (for MHA) reduction in operations, resulting in significantly faster model inference.






Identifying unimportant tokens to be discarded is a major challenge in token pruning. Current methods address this
by either using self-attention values to assign a score to each token, or by adding a prediction module to predict scores. However, both methods have their limitations, which will be discussed in detail.

\subsection{Limitations of attention value-based methods}
\label{sec:analysis1}
\vspace{-1ex}
\begin{figure}[t]
	\centering
	\includegraphics[width=1\columnwidth]{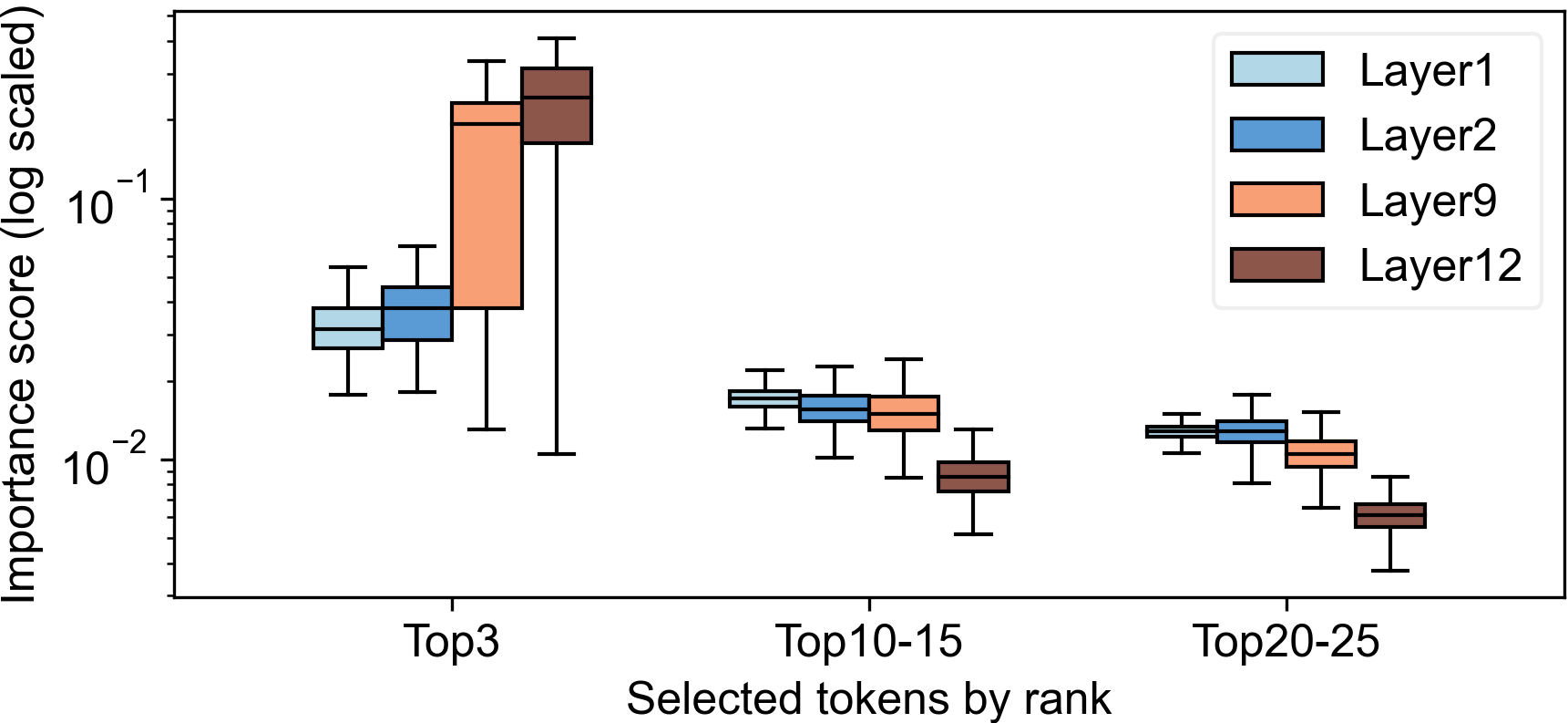}	
	\caption{Comparison of token importance score distributions on the GLUE MRPC dataset using attention values from the BERT$_{base}$ model. Note that the y-axis is log-scaled to better visualization of tokens with low importance scores.}
	\label{fig:attentionscore}
\end{figure}
\vspace{3px}
\noindent\textbf{Token importance scoring}. Attention value-based methods~\cite{powerbert, ltp, spatten} define the importance score of token $x_i$ in layer $l$ as: 
\begin{equation}
	\label{eq:importance_score}
	\begin{aligned}
		\displaystyle  s^l(x_i)=\frac{1}{N_h}\frac{1}{n}\sum_{h=1}^{N_h}\sum_{j=1}^n\bm{A}^l_{h}[x_i, x_j]
	\end{aligned} 
\end{equation}
where $N_h$ denotes the number of heads.  $\bm{A}^l_{h}[x_i, x_j]$ indicates the attention received by token $x_j$ from $x_i$ on head $h$. Thus, token $x_i$ is considered important if it receives more attention from all tokens across all heads. The term $\bm{A}^l_{h}[x_i, x_j]$ can reuse the results from the self-attention mechanism in Equation~\ref{eq:attention}. Therefore, attention value-based scoring in Equation~\ref{eq:importance_score} is computationally lightweight.

However, the attention values scoring in Equation~\ref{eq:importance_score} can inaccurately measure token importance in early transformer layers. 
Previous works~\cite{clark-etal-2019-bert,rogers-etal-2020-primer} have shown that some important tokens, such as [\texttt{SEP}], receive little attention in early layers but gain increased attention in deeper layers. As such, these critical tokens obtain low importance scores in Equation~\ref{eq:importance_score} initially and higher scores in deeper layers.
However, this creates a problem where crucial tokens may be misclassified as unimportant in early layers and removed before they have the chance to reach deeper layers, where they would have received the correct importance scores.

To validate this, we analyze importance score distribution in a trained BERT$_{base}$ model finetuned with the GLUE MRPC dataset. Fig.~\ref{fig:attentionscore} compares the distribution of importance scores at layers 1, 2, 9, and 12. Tokens are selected based on their ranking positions according to the importance scores. The scores of  top3 tokens in deep layers, such as layers 9 and 12, are significantly higher than those of less important tokens ranked in the top10-15 and top20-25. 
However, at early layers such as layers 1 and 2, the scores for top3 tokens  are not well distinguishable from those of other low-ranked tokens. This indicates that early layers tend to assign relatively similar  scores to both top-ranked and low-ranked tokens, potentially resulting in important tokens being deemed unimportant. As a result, while being computationally efficient, using attention value-based scoring can be problematic at early layers.

\subsection{Limitations of prediction-based methods}



 In contrast to the attention value-based scoring method, prediction-based methods~\cite{transkimmer,trbert,dynamicvit} incorporate an additional neural network to predict the importance score for each token. The recently proposed Transkimmer~\cite{transkimmer}  adds an extra prediction module before each transformer layer, which  composes of 2 linear layers with a layernorm~\cite{layernorm}, GeLU activation~\cite{gelu} and GumbelSoftmax~\cite{jang2017categorical}.  These prediction modules gradually update their parameters and learn to predict which tokens should be pruned, which has been shown to outperform attention value-based approaches.

 However, prediction-based token pruning faces limitations in achieving real  latency reduction. First, the  prediction module itself introduces additional inference latency and FLOPs. As shown in Table~\ref{tbl:costcompare}, the additional FLOPs and latency introduced by Transkimmer prediction modules account for 8.43\% and $\sim$30\% of BERT, respectively. This suggests that token pruning needs to  prune much more tokens to counteract the 30\% latency slowdown. 
 Second, Transkimmer's dynamic token pruning relies on the computationally-intensive GumbelSoftmax operation, necessitating specialized runtime and hardware support for efficient implementation.

 \label{sec:analysis2}
 \begin{table}[t]
 	\begin{center}
 		\small
 		\begin{tabular}	{@{\hskip0pt}c@{\hskip2pt}|c@{\hskip2pt}|@{\hskip2pt}c@{\hskip2pt}|@{\hskip2pt}c@{\hskip0pt}}
 			\hline
 			Model &FLOPs &Intel CPU Latency&A100 GPU Latency\\
 			\hline
 			BERT$_{base}$ (12L-768H) &+8.43\%&+11.82\% &+29.35\% \\
 			BERT$_6$ (6L-512H) &+8.48\%& +12.95\%&+28.79\%\\
 			BERT$_4$ (4L-256H) & +8.63\%& +23.53\%&+30.50\%\\
 			\hline
 		\end{tabular}
 		\caption{ Compared to the original model ( input token length=256),
 			Transkimmer~\cite{transkimmer} introduces significant inference latency on both CPU and GPU due to prediction module scoring. FLOPs are calculated using thop~\cite{thop}, while latency measurements are obtained using onnxruntime~\cite{onnxruntime}. 
 		}
 		\label{tbl:costcompare}
 	\end{center}
 \end{table}

In our work, our goal is to implement token pruning in practical applications that accelerate real inference latency. While prediction-based approaches can be challenging in achieving actual latency reduction, 
we instead leverage attention values for token pruning. 

\section{Methodology}

In this section, we present {\sysname}, a novel token pruning approach that incorporates two key techniques for learning the optimal token pruning decisions. First, we introduce the end-to-end token pruning algorithm that leverages $L_0$ regularization. Then, we describe the ranking-aware token distillation approach that enhances the ability of self-attention values to rank token importance. 

\vspace{-1ex}
\subsection{Constraint-aware token pruning}
\vspace{-0.5ex}
\label{sec:tokenpruning}
For a given deployment constraint, we propose constraint-aware token pruning to remove a set of unimportant tokens so that the model with the retained tokens can achieve the best accuracy under the constraint. 
The basic idea is \textit{(1) we introduce a set of binary decision masks $M\in\{0, 1\}$ to represent the sparsity ratio  and indicate whether to drop ($M=0$) or keep each token ($M=1$); (2) use these masks to construct a constraint-aware loss function; (3) optimize the constraint-aware loss using an  improved $L_0$ regularization~\cite{l0} method. } Next, we will introduce the details. 

\begin{figure}[t]
	\centering
	\includegraphics[width=1\columnwidth]{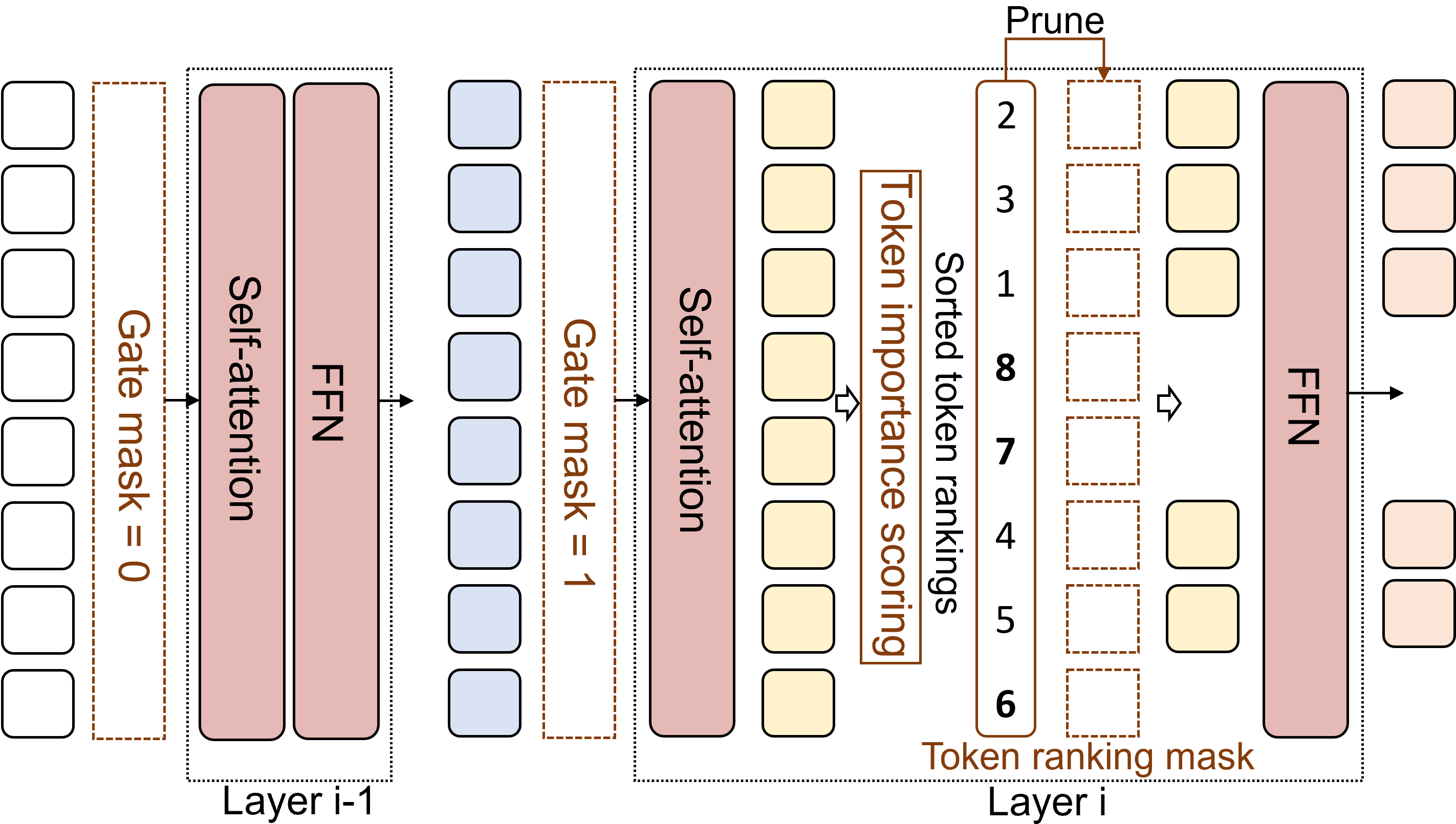}	
	\caption{Our approach learns  layer gate masks and token ranking masks to prune tokens  under a desired constraint. When a layer gate turns off (i.e., mask=0), we skip the current layer. When a layer gate turns on (i.e., mask=1), unimportant tokens are removed after the self-attention mechanism.}
	\label{fig:overview}
\end{figure}

Unlike prior works~\cite{spatten,powerbert,ltp,LAT,trbert,transkimmer} that treat all layers in the same manner, leading to a vast design space, we introduce a novel coarse-to-fine token pruning scheme, as shown in Fig.~\ref{fig:overview}.
 Specifically, gate masks are used to select  a subset of layers for  token pruning, while token ranking masks dynamically determine which specific tokens within the selected layers should be pruned.

 \vspace{3px}
\noindent\textbf{Layer gate mask $M_{gate}$}. We introduce a gate mask $M_{gate}$ for each transformer layer to control whether token pruning is performed in that layer. Concretely, if $M_{gate}^i$ is set to 1, token pruning will be applied to layer $i$. If $M_{gate}^i$ is 0, layer $i$ will be skipped, and retain the 
 tokens from its previous layer $i-1$.

The design choice is built on two insights. First,  it has been observed that pre-trained transformer models have varying levels of token redundancy across different layers~\cite{powerbert,rogers-etal-2020-primer}. Second, pruning tokens across all layers significantly expands the design space and  poses unnecessary difficulty in optimization, particularly  when trying to maintain a low or medium level of compression sparsity. 

In our experiments, we observe an interesting trend where as the target sparsity increases (i.e., constraint becomes tighter), the gate masks progressively activate earlier layers, starting from the deepest one. Eventually, the gate masks activate all transformer layers under high pruning sparsity. Inspired by the observation, we gradually increase the sparsity level from 0 to the target  during  pruning. By doing this, gate masks provide improved decision-making for pruning fewer tokens within early layers. This is because gate masks prioritize learning masks in deeper layers, where token importance is more easily predictable by self-attention values.

\vspace{3px}
 \noindent\textbf{Token ranking position mask $M_{rank}$}. For fine-grained token pruning,  assigning a mask to each token and removing those with a mask value of 0 may seem an intuitive solution. However, it can lead to a problem known as "static token pruning." This occurs because the final mask values are fixed after training, causing  tokens to be pruned at the same positions for all input sequences in the dataset  during inference. This can be problematic as informative tokens can appear in different positions across different input sequences. Removing tokens  at the same positions for all input sequences can also inadvertently remove important tokens and result in a significant loss of accuracy. 
 


To address this challenge, we follow PoWER-BERT~\cite{powerbert} to use ranking masks (see Fig.~\ref{fig:overview}). Instead of applying a mask to each token directly, we mask tokens'  ranking positions  based on their importance score, which is  computed by utilizing attention values, as outlined in Equation~\ref{eq:importance_score}. The insight is that by scoring tokens based on their importance to the final model prediction, crucial tokens will always rank in the topmost positions after sorting. This means that although informative tokens may appear in different positions based on input content, their ranking positions are static (i.e., topmost) and can be indicated by a static mask. For example, given  an input sequence of $X_{i-1}=(x_1, x_2,...x_n)$ for layer $i$, we sort the tokens by their importance scores, resulting in a ranking of  $(n, 3,...1)$. The corresponding ranking masks are then defined as $(M_{rank}^{i,n}, M_{rank}^{i,3},...M_{rank}^{i,1})$, where the value of $M_{rank}^{i,j}$ indicates whether prune or keep the $j^{th}$ ranked tokens for layer $i$.

\vspace{3px}
\noindent\textbf{FLOPs-aware constraint}. With these masks, we can calculate the number of retained tokens and use them to measure the computation and memory cost that is required for  model inference. In our work, we use FLOPs as a metric for evaluating the cost of model inference due to its ease of use. Formally, let $\bm{M}=\{M_{gate}^1,...M_{gate}^L, M_{rank}^{1,1},...\}$,  denoting all the inserted masks. Then the expected model FLOPs after token pruning can be calculated from $\bm{M}$ as follows:
\begin{equation}
	\label{eq:sparsity}
	\begin{aligned}
		\displaystyle  c(\bm{M})=\Sigma_{i=1}^{L}(4\cdot \text{d}^2\cdot T_i+2\cdot \text{d}\cdot T_i^2+N_{h}\cdot T_i^2)\\+\Sigma_{i=1}^{L}(2\cdot \text{d}\cdot d'\cdot T_i)
	\end{aligned} 
\end{equation}
where the two items calculate the FLOPs of MHA and FFN layers, respectively.   $d$ denotes the hidden size, $N_h$ is the number of heads in MHA layer, $d'$ denotes FFN intermediate size.  $T_i$ represents the number of tokens that are retained for  the $i^{th}$ layer, which can be easily computed by multiplying $\mathbb{E}(M_{rank}^i>0)$ with the original token length $n$. Note that when the gate mask $M_{gate}^i$ turns off fine-grained token pruning, $T_i$ keeps the same number of tokens with its previous layer (i.e., $T_i$ =$T_{i-1}$).


\vspace{3px}
\noindent\textbf{Learning masks under a desired constraint}.  Now we introduce how to determine the values of gate masks and ranking masks for minimal accuracy loss under a given FLOPs constraint. 
Let $\bm{\theta}$ denote the original model and $\bm{M}$  denote all the pruning masks.  We formalize the task of token pruning as an end-to-end learning problem by adding a regularization term $L_{reg}$:

\begin{equation}
	\label{eq:loss1}
	\begin{aligned}
		L=L_{downstream}(\bm{\theta},\bm{M})+\lambda L_{reg} (\bm{M})
	\end{aligned} 
\end{equation}

Optimizing the above loss function requires the masks $M_{gate}$ and $M_{rank}$ to be differentiable. However, the original masks are discrete binary values.
To overcome this, we use the $L_0$ reparameterization method proposed by ~\cite{l0}, which is specifically designed for model pruning. Following the standard $L_0$ reparameterization, masks $\bm{M}$ are regulated using the hard concrete distribution as follows
\begin{equation}
	\label{eq:l0}
	\begin{aligned}
	\bm{u}\sim U(0,1)\\
	\text{s}=\text{sigmoid}((\text{log}\frac{\bm{u}}{1-\bm{u}}+\text{log}\bm{\alpha)}/\beta)\\
	\tilde{s}=s\times(r-l)+l\\
	\bm{M}=\text{min(1, max}(0,	\tilde{s}))
	\end{aligned} 
\end{equation}
where $U(0,1)$ is a uniform distribution in the interval [0,1]; $l<0$ and $r>0$ are two constants that stretch the sigmoid output into the interval $(l,r)$. $\beta$ is  a hyperparameter that controls the steepness of the sigmoid function. We adopt the common practice of setting $l$ to -0.1, $r$ to 1.1 and $\beta$ to $\frac{2}{3}$.  $\bm{\alpha}=\{\alpha_j\}_{j=1}^{|\bm{M}|}$ are the main learnable parameters. 
The hard concrete distribution assigns a significant portion of probability mass on the integer values \{0,1\}, which serves as a good continuous approximation of the binary (Bernoulli) distribution. 

During training, the hard concrete parameters $\bm{\alpha}$ and $\bm{u}$ determine the values of masks $\bm{M}$. We learn masks $\bm{M}$ by updating these learnable parameters of the distributions
from which the masks are sampled in the forward pass.
Moreover, these learnable parameters and masks can be jointly optimized with the original model parameters, resulting in better performance.

In prior token pruning works~\cite{powerbert,transkimmer,ltp}, $L_1$  is widely used as the regularization loss in Equation~\ref{eq:loss1}, and $\lambda$ controls the trade-off between final model accuracy and the achieved sparsity. However, it requires careful hyper-parameter tuning to make sure it converges to a desired sparsity~\cite{cofi,lagrangian},  which lacks effective control on the complexity of final model inference.

In our work, we aim to control the achieved model FLOPs after token pruning. We  follow~\cite{lagrangian} to replace the vanilla $L_0$ objective with a Lagrangian multiplier. 
Let $C$ be the target FLOPs, $c(\bm{M})$ be the expected FLOPs determined by the masks $\bm{M}$ in Equation~\ref{eq:sparsity}.   We  impose an equality constraint $c(\bm{M})=C$ by introducing a  penalty:
\begin{equation}
	\label{eq:loss3}
	\begin{aligned}
		\displaystyle L_{reg}(\bm{M})=\lambda_1\cdot(c( \bm{M})-C)+\lambda_2\cdot(c(\bm{M})-C)^2
	\end{aligned} 
\end{equation}
where the masks $\bm{M}$ are determined by hard concrete parameters $\bm{\alpha}$ and $\bm{u}$ in Equation~\ref{eq:l0}.



\vspace{-1ex}
\subsection{Distillation of token importance rankings}
\label{sec:distillation}

In our proposed constraint-aware token pruning algorithm, the ranking masks are applied to the sorted tokens based on their importance scores.  Ideal importance scoring mechanism that provides reliable token importance rankings is essential for maintaining the performance of the final model prediction. The most crucial tokens will be at the top and retained by the ranking masks.

In {\sysname}, we utilize self-attention values to calculate token importance (Equation~\ref{eq:importance_score}) for ease of hardware deployment. However, as discussed in Sec.~\ref{sec:analysis1}, accurately ranking the importance of tokens using the self-attention mechanism is challenging in the early  layers of the transformer. Despite the potential for improved token rankings in deeper layers, crucial tokens may be mistakenly ranked as unimportant and removed before they have the opportunity to reach these layers, leading to a substantial decrease in accuracy.

To overcome this limitation, we propose a method to distill more accurate ranking decisions to the early layers during the training process, enhancing the ability of self-attention values to rank token importance. To this end, we enable the preservation of the most important tokens during inference, thus improving overall accuracy.


\begin{figure}[t]
	\centering
	\includegraphics[width=0.95\columnwidth]{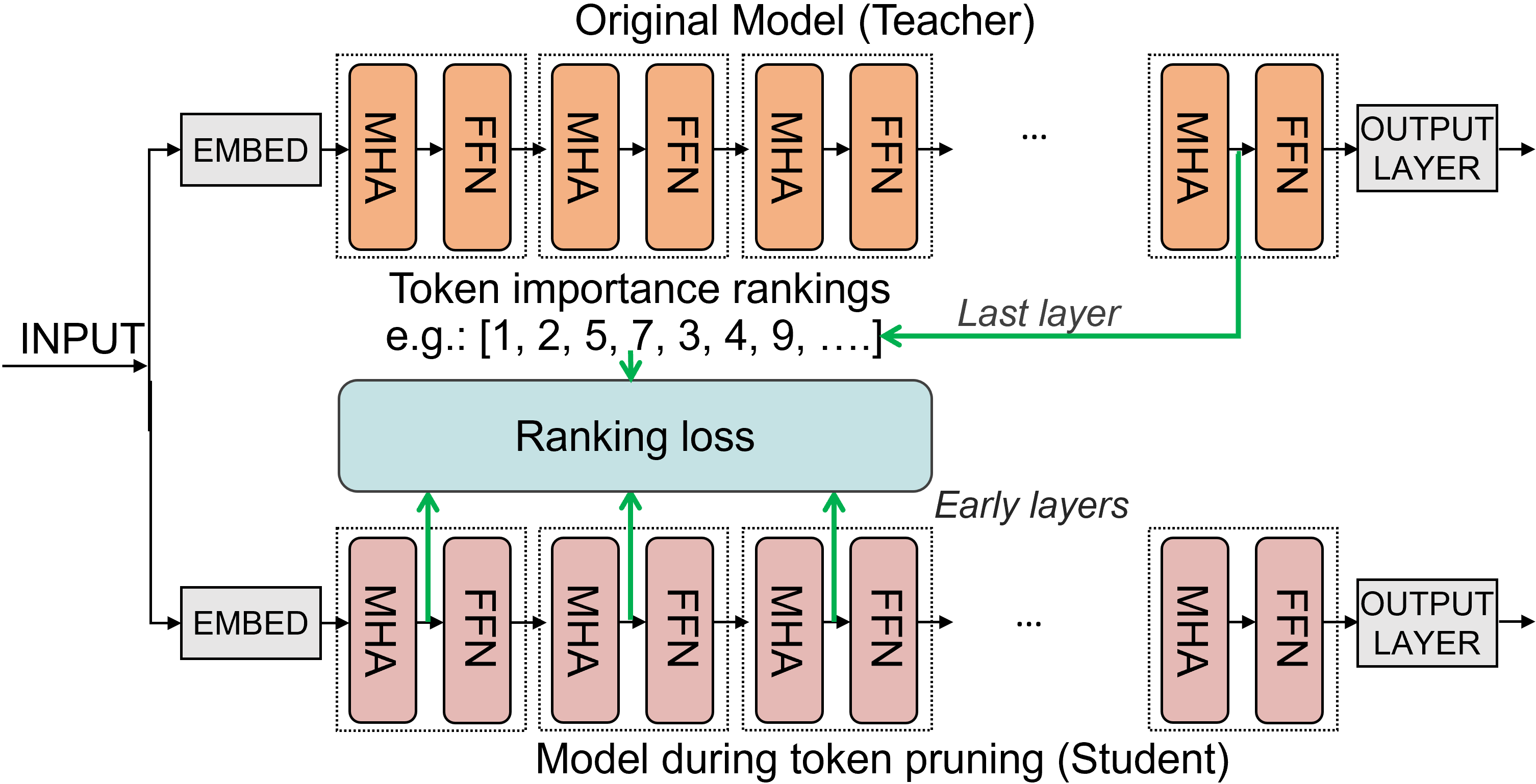}
	\vspace{-2ex}	
	\caption{Our ranking-aware token distillation uses  importance rankings generated from the unpruned model's final layer and distill it to early layers during the training.}
	\label{fig:kd}
\end{figure}

\vspace{3px}
\noindent\textbf{Ranking-aware token distillation}. Before distillation, we need to obtain accurate token importance rankings as the knowledge. However, obtaining such ground truth or labeled rankings is challenging. Our experiments, as shown in Fig.~\ref{fig:attentionscore}, suggest that attention values in final layers can effectively identify important tokens. Thus, we use the token importance rankings generated from the final transformer layer of the unpruned model as our knowledge source. 

The next challenge is to develop an effective distillation objective. Previous studies~\cite{movement, cofi, nn_pruning}  have shown that combining distillation with model weight pruning can improve performance. These studies have used distillation objectives based on cross-entropy loss~\cite{kd, sanh2020distilbert} or Mean Squared Error (MSE) loss between the student and teacher layer-wise representations~\cite{cofi, Sun2019PatientKD, jiao2020tinybert}.
However, when it comes to token distillation, the conventional MSE loss may not be suitable due to the different levels of information captured by early and deep layers of transformer models like BERT~\cite{jawahar-etal-2019-bert}. The representations of early and deep layers are meant to be distinct, making it difficult to use conventional distillation objectives.


Instead, we  use token importance rankings from the final layer of the teacher model as the information to be distilled and develop a ranking-aware distillation loss. Our objective is to align the token importance rankings of the early transformer layers with those of the final layer in the teacher model. This way, self-attention layers are encouraged to give more importance to the crucial tokens, increasing the chance of retaining them in early layers. 

Fig.~\ref{fig:kd} illustrates the overview process of our proposed ranking-aware token distillation. To effectively distill the knowledge, we adopt a ranking loss as the distillation objective. Specifically, we use the LambdaLoss  as proposed in~\cite{rankloss} to optimize Normalized Discounted Cumulative Gain (NDCG), which is a widely used ranking metric in information retrieval systems that places more emphasis on the importance of top-ranked tokens.

 Specifically, let $\bm{R}$ denote the rankings obtained by sorting the importance scores of each token in the final layer in the teacher model.  $\bm{S}^i$ refers to the token importance scores in the $i^{th}$ layer of the student model, and $\bm{X}$ represents a mini-batch of training data. We define the distillation loss as follows: 
	\begin{equation}
	\label{eq:distill}
	\begin{aligned}
		\displaystyle  
		L_{distill}=\sum_{i=1}^{L'} LambdaLoss(\bm{R},\bm{S}^i, \bm{X})
	\end{aligned} 
\end{equation}

 In our experiments, we empirically set the first 1/3 layers ($L'=\frac{1}{3}L$) as the early layers for distillation.

\subsection{Training and inference}
We now describe the full training process of our {\sysname}. Given a deploy constraint, {\sysname} simultaneously trains  the masks and model parameters, allowing for effective token pruning  and improved model adaptation to token sparsification. Additionally, we incorporate ranking-aware distillation to enhance the ability of the attention values in determining the importance of tokens in early layers.
The full training objective is a combination of the above objectives:
\begin{equation}
	\label{eq:finalloss}
	\begin{aligned}
		L=L_{downstream}(\bm{\theta}, \bm{M}) + L_{reg}(\bm{M}) + \lambda L_{distill}
	\end{aligned} 
\end{equation}
where $L_{reg}(\bm{M})$ is defined as in Equation~\ref{eq:loss3}, and $\lambda$ is a hyperparameter controlling the contribution of distillation loss. In particular, the two hyperparameters $\lambda_1$ and $\lambda_2$ introduced by $L_{reg}(\bm{M})$ are automatically adjusted using the AdamW optimizer. Therefore, the only additional hyperparameter that needs to be tuned is $\lambda$.

Once the training finishes, the token pruning decisions are determined by combining the gate masks and ranking masks. Specifically, only the layers chosen by the gate masks with $M_{rank}=1$ are assigned with ranking masks for token selection. The other layers are not considered for token pruning.

During the inference, we use self-attention values to calculate token importance scores in the selected layers. We then sort the tokens based on these scores.
 By discarding the tokens through use of ranking masks ($M=0$), we can effectively eliminate unnecessary tokens and improve the efficiency of the model.

\begin{table*}[t]
	\begin{center}
		\small
		\begin{tabular}	
			{@{\hskip0pt}c@{\hskip2pt}c@{\hskip5pt}c@{\hskip5pt}c@{\hskip5pt}c@{\hskip5pt}c@{\hskip5pt}c@{\hskip5pt}c@{\hskip5pt}c@{\hskip5pt}c@{\hskip5pt}c@{\hskip5pt}c@{\hskip5pt}c@{\hskip5pt}c@{\hskip5pt}c@{\hskip5pt}c@{\hskip5pt}c@{\hskip5pt}c@{\hskip5pt}c@{\hskip5pt}c@{\hskip0pt}}
			\hline
			\multicolumn{18}{c@{\hskip0pt}}{\textbf{(a)  Token pruning on BERT$_{base}$}}\\
			\hline
			\multirow{2}{*}{Model} & 	\multirow{2}{*}{Method} &  \multicolumn{2}{c}{CoLA} &  \multicolumn{2}{c}{RTE}&  \multicolumn{2}{c}{QQP}&  \multicolumn{2}{c}{MRPC}&  \multicolumn{2}{c}{SST-2}&  \multicolumn{2}{c}{MNLI}&   \multicolumn{2}{c}{QNLI}&  \multicolumn{2}{c@{\hskip0pt}}{STS-B}\\        
			&&Matthews &FLOPs&Acc. & FLOPs&Acc. & FLOPs & F1 & FLOPs & Acc. & FLOPs & Acc. & FLOPs & Acc. & FLOPs & Pearson & FLOPs\\
			\midrule
			BERT$_{base}$ & -	& 57.8 &  1.00$\times$ & 65.7&  1.00$\times$ & 91.3&  1.00$\times$ & 88.9 &  1.00$\times$ & 93.0 &  1.00$\times$ & 84.9 &  1.00$\times$ & 91.4 &  1.00$\times$ & 88.6&  1.00$\times$ \\
			\hline
			PoWER-BERT& Atten-value& 52.3 &  4.50$\times$ & 67.4&  3.40$\times$ & 90.2&  4.50$\times$& 88.1 &  2.70$\times$ & 92.1 &  2.40$\times$ & 83.8 &  2.60$\times$ & 90.1 &  2.00$\times$ & 85.1&  2.00$\times$ \\
			LAT & Atten-value  &- &  - & -&  - & -& - &- &  - & 92.8 &  2.90$\times$ & 84.4 &  2.80$\times$ & -& -& -&  -\\
			LTP & Atten-value & 52.3 & 8.66$\times$ & 63.2 & 6.84$\times$ & 90.4 & 7.44$\times$ & 87.1 & 6.02$\times$ & 92.3 & 3.59$\times$ & 83.9 & 3.74$\times$  & 89.3 & 3.91$\times$ & 87.5 & 5.25$\times$ \\
			\textbf{\sysname} & Atten-value&\textbf{60.5}& \textbf{9.62$\times$} & \textbf{70.0} & \textbf{7.10$\times$} & \textbf{91.2} & \textbf{8.04$\times$} &\textbf{89.2} & \textbf{6.32$\times$} &  \textbf{93.5}& \textbf{3.82$\times$} &\textbf{84.7}& \textbf{4.27$\times$} & \textbf{90.6} & \textbf{4.35$\times$} & \textbf{87.8} & \textbf{5.32$\times$} \\
			\hline
			Transkimmer& Prediction  &  58.9 &  9.81$\times$ & 68.9&  11.22$\times$ & 90.8&  11.72$\times$ & 88.5 &  7.45$\times$ & 92.3 &  4.07$\times$ & 83.2 &  6.65$\times$ & 90.5&  6.01$\times$ & 87.4&  7.24$\times$ \\
			\textbf{\sysname} & Atten-value& \textbf{60.0}& \textbf{10.05$\times$} &67.9 & 9.94$\times$ &\textbf{90.9}& \textbf{12.57$\times$} &\textbf{88.8} & \textbf{7.47$\times$} & \textbf{93.2} & \textbf{4.52$\times$} &\textbf{83.2}& \textbf{6.65$\times$} &89.1 & \textbf{6.14$\times$} & 86.4 & \textbf{7.30$\times$} \\
			\hline 
				\multicolumn{18}{c@{\hskip0pt}}{\textbf{(b)  Token pruning on RoBERTa$_{base}$}}\\
			\hline
			RoBERTa$_{base}$ & - & 61.8 &1.00$\times$ &78.0 &1.00$\times$&90.4 &1.00$\times$& 92.1 &1.00$\times$& 94.3 &1.00$\times$& 87.5 &1.00$\times$&92.9 &1.00$\times$&90.9 &1.00$\times$\\
			\hline
			LTP &Atten-value & - & - & 78.0 &6.93$\times$& 89.7 &8.70$\times$& 91.6 &4.99$\times$& 93.5 &5.28$\times$& 86.5&6.09$\times$& 92.0 &4.61$\times$& 90.0&3.81$\times$\\
			Transkimmer &Prediction & 61.3 & 8.55$\times$ & 76.2 &6.74$\times$&91.0  &19.40$\times$& 91.9 &6.22$\times$& 93.5 &5.26$\times$& 86.7&7.01$\times$&  91.7&7.00$\times$& 90.5&5.23$\times$\\
			{\sysname} &Atten-value & \textbf{64.5} & \textbf{8.91$\times$} &\textbf{79.4} &\textbf{7.16$\times$}&90.6  &\textbf{19.42$\times$}& \textbf{92.3} &\textbf{6.38$\times$}& \textbf{93.8} &\textbf{5.56$\times$}& \textbf{86.9}&\textbf{7.04$\times$}& \textbf{92.8} &5.02$\times$&89.6 &4.02$\times$\\
			
			\hline
		\end{tabular}
		\caption{ Performance and FLOPs reduction compared to state-of-the-art token pruning methods on GLUE benchmark.  }
		\label{tbl:results1}
	\end{center}
\end{table*}

\section{Experiments}
\subsection{Experiment Setup}
\noindent\textbf{Datasets and Models}. 
We   evaluate {\sysname} on a diverse set of datasets, including 8 classification and regression tasks from the GLUE benchmark~\cite{glue}, the SQuAD-v2.0~\cite{squadv2} extractive question answering dataset, and the 20News~\cite{20news} long sequence classification dataset. The diversity of these tasks and the range of token lengths (i.e., 64, 128, 256, 384, 512) showcase the general applicability of our method. A detailed summary of these datasets can be found in Appendix. To demonstrate the generality of our approach on both large and small pre-trained language models, we implement {\sysname} on RoBERTa$_{base}$~\cite{roberta}, BERT$_{base}$ and BERT$_6$.

\vspace{2pt}
\noindent\textbf{Training setup}. Following existing works~\cite{powerbert,ltp,transkimmer}, we perform  fine-tuning on the downstream datasets before the token pruning. Then we conduct token pruning with the optimization objective in Equation~\ref{eq:finalloss} under multiple  FLOPs constraints. We use \texttt{{FLOPs sparsity}} to denote the constraint, which is defined as the ratio of FLOPs that have been pruned or removed from the model.  We also define \texttt{FLOPs reduction} as the ratio of the full model FLOPs under the original input sequence length to the remaining FLOPs.

 The token pruning process begins with a warmup stage where we gradually increase sparsity to the target value using a linear scheduler. The initial value of $\lambda$ is determined by hyper-parameter grid search from [1e-2, 1e-3, 1e-4] on development set with 20\% data randomly picked from training set.   At the start, layer gate masks are initialized to 0, and ranking masks are initialized to 1, meaning all tokens are retained. After reaching the final sparsity level, we fix it and continue training the model and pruning masks until convergence (i.e., fine-tuning stage). See Appendix for detailed settings of other hyper-parameters.  We conduct all  experiments
 with random seed 57.


\vspace{2px}
\noindent\textbf{Baselines}. We compare {\sysname} with the state-of-the-art token pruning and model compression methods. Token pruning baselines include attention value-based methods such as PoWER-BERT~\cite{powerbert}, LAT~\cite{LAT}, LTP~\cite{ltp}, and strong prediction-based method  Transkimmer~\cite{transkimmer}. Model compression baselines include DistilBERT~\cite{sanh2020distilbert}, and CoFi~\cite{cofi}, 
which is  a state-of-the-art structured pruning method.

 \begin{table}[t]
 	\begin{center}
 		\small
 		\begin{tabular}	
 			{cccccc}
 			\hline
 			\multirow{2}{*}{Model} & \multirow{2}{*}{Method}  &  \multicolumn{2}{c}{SQuADv2.0} &\multicolumn{2}{c}{20News} \\
 			& &F1&FLOPs&Acc. & FLOPs \\
 			\midrule
 			BERT$_{base}$ & - & 77.1 & 1.00$\times$ & 86.7&1.00$\times$ \\
 			
 			Transkimmer &Prediction&75.7 & 4.67$\times$ & 86.1&$ 8.11\times$ \\
 			PoWER-BERT& Atten-value& - & - & 86.5 & 2.91$\times$ \\
 			LTP& Atten-value& 75.6 & 3.10$\times$ & 85.2 & 4.66$\times$ \\
 			\textbf{{\sysname}}&Atten-value & \textbf{75.9} & \textbf{4.12$\times$} & \textbf{87.0}&\textbf{8.26$\times$}  \\
 			\hline
 		\end{tabular}
 		\caption{ Results by different token pruning methods on downstream tasks with long sequence length.}
 		\label{tbl:results2}
 	\end{center}
 \end{table}

			

\begin{figure*}[ht]
	\centering
	\includegraphics[width=1\textwidth]{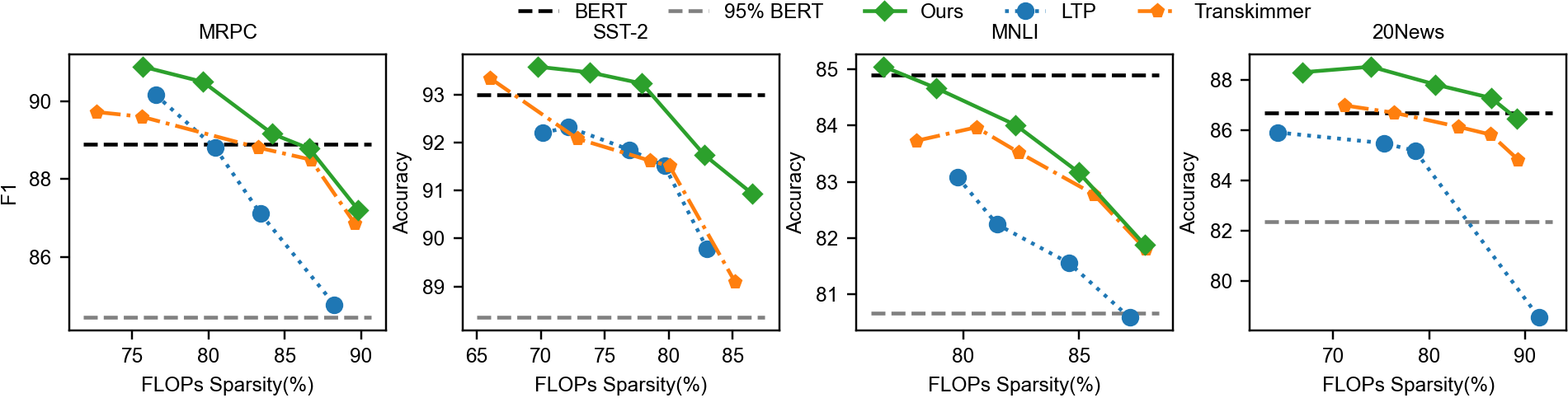}	
	\vspace{-5ex}
	\caption{Comparison of token pruning methods under various FLOPs sparsity ratios.}
	\label{fig:sparsity}
\end{figure*}

\subsection{Main results}
\textbf{Overall performance}. We evaluate  {\sysname} on both BERT$_{base}$ and RoBERTa$_{base}$, and compare with the state-of-the-art token pruning baselines.  LTP lacks an implementation for BERT$_{base}$, we  follow their official code and implement for BERT$_{base}$. For a fair comparison, we follow the provided instructions and conduct a grid search for the optimal hyper-parameters. To prove {\sysname}'s effectiveness on varying input sequence lengths, we conduct experiments on both the GLUE benchmark and two long sequence tasks.

We start by comparing {\sysname} to the original BERT and RoBERTa. As shown in Table~\ref{tbl:results1}  and Table~\ref{tbl:results2}, {\sysname} achieves comparable accuracy on GLUE benchmark, SQuAD and 20News while significantly reducing FLOPs. For instance,  {\sysname} even outperforms the original BERT with +2.7\%, +4.3\%, +0.3\%, and +0.5\% improvement on CoLA, RTE, MRPC, and SST-2 respectively, and achieves an average FLOP reduction of 6.7$\times$.   On other datasets, the accuracy drop is minimal at $<0.8\%$. 
These results demonstrate the effectiveness of {\sysname} in reducing computational cost while maintaining accuracy.

Compared to other token pruning methods based on attention values,  {\sysname} significantly surpasses strong baselines such as PoWER-BERT, LAT, and LTP across all datasets, despite being based on the same approach. This is mainly due to our ranking-aware token distillation mechanism, which greatly enhances the ability of self-attention values to rank token importance.
Also, as can be seen from Table~\ref{tbl:results1} and Table~\ref{tbl:results2}, current attention value-based methods cannot surpass the prediction method-based Transkimmer. However,  {\sysname} outperforms Transkimmer on many tasks. Specifically, under the same-level FLOPs reduction, {\sysname} on BERT achieves +2.9\%, +0.1\%, +0.3\%, +0.9\%, +0.2\%, +0.9\% higher accuracy on CoLA, QQP, MRPC, SST-2, MNLI, SQuAD  and 20News, respectively. 

It's worth noting that {\sysname} also outperforms Transkimmer in terms of real inference latency improvement. As discussed in Sec.~\ref{sec:analysis2}, it's challenging to deploy Transkimmer for real latency reduction. In contrast, {\sysname} delivers real latency improvement of up to 7.4$\times$, which will be further discussed in later sections.


\noindent\textbf{Under various deployment constraints.} 
We now evaluate the effectiveness of {\sysname} under various FLOPs sparsity and compare it to two state-of-the-art methods: LTP (representing attention value-based methods) and Transkimmer (representing prediction-based methods). For a fair comparison, we use their official code implementations and perform a grid search to find the optimal hyper-parameters that achieve the desired token sparsity.

Fig.~\ref{fig:sparsity} summarizes the results.  {\sysname} surpasses both LTP and Transkimmer at all FLOPs sparsity levels.
At equivalent levels of accuracy on the 20News long sequence, {\sysname} reduces 17\% and 20\% more FLOPs compared to Transkimmer and LTP respectively. 
 Without any loss in BERT$_{base}$ accuracy, {\sysname} reduces FLOPs by 86\%, 78\%, 76\%, and 88\% on MRPC, SST-2, MNLI, and 20News, respectively.

\noindent\textbf{Comparison with model compression}. In addition to token pruning baselines, we compare {\sysname} with state-of-the-art model compression techniques, including structured pruning and knowledge distillation (i.e., the DistilBERT$_6$). 
We specifically compare with CoFi~\cite{cofi}, a  top-performing structured pruning method. We evaluate CoFi and {\sysname} on two model sizes: BERT$_{base}$, a large transformer model, and BERT$_6$, a small model.

Table~\ref{tbl:compression1} shows the results by different methods. {\sysname} and CoFi outperform DistilBERT$_6$ in terms of higher compression ratios. {\sysname} consistently surpasses the  original BERT$_{base}$ consistently with higher accuracy and a 6.7$\times$ average FLOPs reduction, whereas CoFi sees a significant drop in accuracy for CoLA and 20News at high compression ratios. 
 On the smaller BERT$_6$ model, {\sysname}  exhibits a much smaller accuracy loss compared to CoFi, showcasing its effectiveness and robustness across models of various sizes.

\begin{table}[t]
	\begin{center}
		\small
		\begin{tabular}	{@{\hskip0pt}c@{\hskip2pt}c@{\hskip4pt}c@{\hskip5pt}c@{\hskip4pt}c@{\hskip5pt}c@{\hskip4pt}c@{\hskip5pt}c@{\hskip4pt}c@{\hskip0pt}}
			\hline
			Model  &\multicolumn{2}{c}{CoLA} &\multicolumn{2}{c}{MRPC} &\multicolumn{2}{c}{MNLI} &\multicolumn{2}{c}{20News} \\
			&Matthews &FLOPs&F1 & FLOPs&Acc. & FLOPs&Acc. & FLOPs\\
			\hline 
			BERT$_{base}$  &57.8 &1.0$\times$ &88.9&1.0$\times$&84.9&1.0$\times$&86.7&1.0$\times$\\
			DistilBERT$_6$ &49.0 & 2.0$\times$&86.9&2.0$\times$&82.6&2.0$\times$&85.8&2.0$\times$\\
			CoFi&39.8&9.1$\times$&90.0&4.0$\times$&84.7&4.0$\times$&86.4&7.7$\times$\\
			\textbf{{\sysname}}&\textbf{60.0}&\textbf{10.0$\times$}&\textbf{90.9}&\textbf{4.1$\times$}&\textbf{85.0}&\textbf{4.3$\times$}&\textbf{87.0}&\textbf{8.2$\times$}\\
			\hline
			\hline
			BERT$_6$  &49.0 &1.0$\times$ &90.4&1.0$\times$&82.6&1.0$\times$&87.3&1.0$\times$\\
			CoFi&38.0&9.1$\times$&86.3&7.7$\times$& 
80.1 & 6.6$\times$&85.9&5.9$\times$\\
			\textbf{{\sysname}}& \textbf{48.9} & \textbf{9.5$\times$}&\textbf{87.8}&\textbf{7.7$\times$}&\textbf{82.0} & \textbf{6.6$\times$}& \textbf{86.8} & \textbf{6.0$\times$}\\
			\hline			
			
		\end{tabular}
		\caption{Comparison with state-of-the-art model compression methods on both large and small pre-trained models.  }
		\label{tbl:compression1}
	\end{center}
\end{table}

\subsection{Ablation study}
\begin{table}[t]
	\begin{center}
		\fontsize{7.5}{7.5} \selectfont
		\begin{tabular}	{@{\hskip0pt}c@{\hskip-3pt}c@{\hskip2pt}c@{\hskip3pt}c@{\hskip3pt}c@{\hskip3pt}c@{\hskip3pt}c@{\hskip3pt}c@{\hskip3pt}c@{\hskip0pt}}
			\toprule
		 \multirow{2}{*}{Method}& CoLA & RTE & QQP & MRPC & SST-2&MNLI&QNLI&STS-B\\
		   &Matthews&Acc.&Acc.&F1&Acc.&Acc.&Acc.&Peason\\
			\toprule
			{\sysname} &\textbf{60.0}&\textbf{67.9}&\textbf{90.9}&\textbf{88.8}&93.2&\textbf{83.2}&\textbf{89.1}&\textbf{86.4}\\
			{\sysname} ${}-$ Gate Masks &58.9 &61.7 &89.4&85.3&\textbf{93.5}&81.7&87.3&83.5\\
			\midrule
			{\sysname} & \textbf{60.0} & \textbf{67.9}&\textbf{90.9}&\textbf{88.8}&\textbf{93.2}&\textbf{83.2}&\textbf{89.1}&\textbf{86.4}\\
			{\sysname} ${}-$ Distill& 59.4 & 60.6 & 90.7&86.4& 92.9&83.0& 88.0& 82.7\\
			{\sysname} (MSE Distill loss) &57.6 &62.8 &90.6&87.5&93.0&82.9&87.9&73.5\\
			{\sysname} (CE Distill loss) &59.0 &60.3 &90.5&88.3&92.3&82.7&88.1&84.8\\
			\hline
		\end{tabular}
		\caption{Alabtion studies of different methods under the same token pruning sparisity. Above: effectiveness of our coarse-to-fine grained pruning strategy. Below: ablation study of different token importance distillation objectives.   }
		\label{tbl:ablationstudy}
	\end{center}
\end{table}

\noindent\textbf{The effectiveness of coarse-to-fine grained pruning strategy}. Compared to prior works, {\sysname} introduces a new coarse-to-fine pruning approach that uses gate masks to choose the best subset of transformer layers for fine-grained token pruning. To evaluate the impact of this design on performance, we compare the results with and without gate masks by removing the masks and pruning all layers at the fine-grained level. We run all experiments under the same FLOPs sparsity (\cref{tbl:results1,tbl:results2}) and hyper-parameter settings.

Table~\ref{tbl:ablationstudy} shows the accuracy results on  GLUE benchmark. Removing gate masks causes a noticeable accuracy drop  on all datasets except SST-2. Specifically, the accuracy drops by a substantial  6.2\%, 3.5\%, 2.9\%, and 2.9\% on RTE, MRPC, QNLI, and STS-B, respectively. These results highlight the effectiveness of our coarse-to-fine grained  approach in making much better token selection decisions.

\vspace{3px}
\noindent\textbf{Different token  distillation approaches}. We also ablate on the ranking-aware token distillation component to evaluate its contribution to the performance of {\sysname}. We first remove the entire distillation process from the token pruning process. As shown in Table~\ref{tbl:ablationstudy}, the removal of our proposed ranking-aware token distillation results in accuracy drops across all datasets. Moreover, we observe  that the effect of ranking-aware token distillation varies based on the length of the input sequences. On datasets with relatively short sample lengths (see Table~\ref{tbl:dataset_summary} in Appendix), such as CoLA, QQP, SST-2, and MNLI, token distillation slightly improves accuracy by 0.6\%, 0.2\%, 0.3\%, and 0.2\%, respectively. However, on datasets with longer sample lengths, such as RTE, MRPC, QNLI, and STS-B, token distillation has a crucial role in improving accuracy by 7.2\%, 4.4\%, 1.1\%, and 3.7\%, respectively. The underlying reason is that it's much easier for conventional attention-based scoring methods to determine the significance of tokens in shorter sequences compared to longer sequences. However, when the task involves longer sequences, accurately identifying the most critical tokens at early transformer layers becomes challenging. In such cases, our proposed ranking-aware token distillation effectively tackles this problem and leads to a significant improvement in accuracy.

Moreover, we  compare the use of conventional distillation objectives to the $LambdaLoss$ in Equation~\ref{eq:distill}. 
Our alternatives include the MSE loss~\cite{cofi} which seeks to reduce the discrepancy between  teacher's and student's token importance scores, and the general cross-entropy (CE) loss~\cite{kd} that aims to minimize the KL divergence between teacher's and student's token importance score distributions. Table~\ref{tbl:ablationstudy}  indicates that relying on conventional distillation objectives fails to improve the accuracy effectively.

\begin{figure}[t]
	\centering
	\includegraphics[width=1\columnwidth]{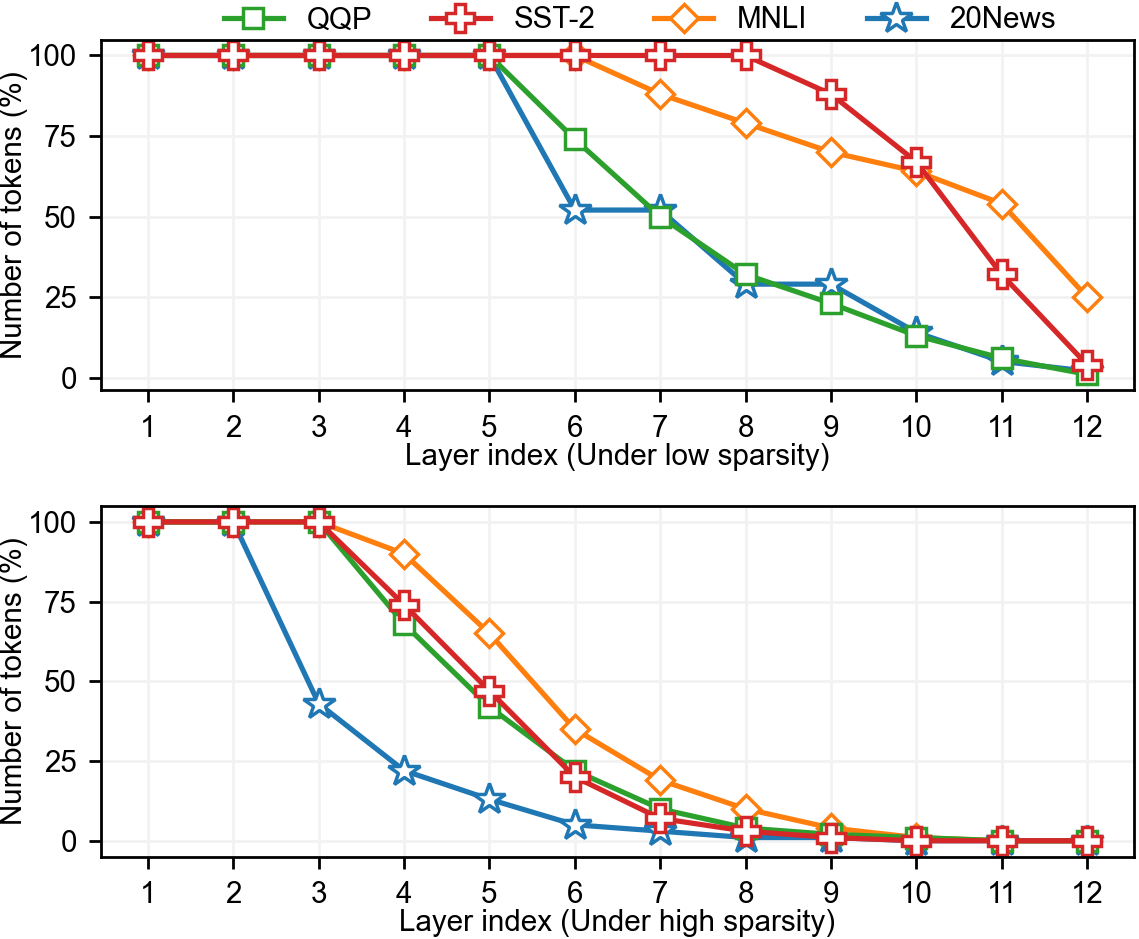}	
	\vspace{-3ex}
	\caption{The retained tokens as input for each layer in BERT. }
	\label{fig:insight}
\end{figure}

\subsection{Retained Tokens Visualization and Analysis}
To further understand the behavior of {\sysname}, we analyze the number of retained tokens in each layer. We conduct experiments on four datasets using two groups of FLOPs sparsity ratios: (i) low sparsity {45\%, 20\%, 20\%, 50\%} and (ii) high sparsity {65\%, 65\%, 60\%, 80\%} for QQP, SST-2, MNLI, and 20News respectively.  The low sparsity numbers are used for experiments in Fig.~\ref{fig:sparsity} and the high sparsity numbers are evaluated in Table~\ref{tbl:results1} and Table~\ref{tbl:results2}. The model accuracy loss  is negligible as demonstrated in previous sections. Notably, for better visualization, we exclude \texttt{PAD} tokens when analyzing {\sysname}'s  pruning decision on the original effective input tokens. 

Fig.~\ref{fig:insight} illustrates the number of remaining tokens used as input for each layer in BERT$_{base}$ under different levels of sparsity ratios. Interestingly, we discover common token pruning patterns in the models: (1) deep transformer layers have high token redundancy. Under high sparsity,   over 95\% of tokens are pruned when they arrive at layer 8, indicating that earlier layers have effectively extracted their information, making them non-informative for deep layers. This observation is consistent with the conclusions of  studies on BERT behaviours~\cite{clark-etal-2019-bert,rogers-etal-2020-primer}, indicating that {\sysname} is capable of automatically identifying the optimal patterns for token pruning. (2) {\sysname} prioritizes token pruning in deeper layers over an even distribution across all layers. We observe that {\sysname} selects different layers for token pruning under varying levels of sparsity ratios. In {\sysname}, token pruning is initially performed in deeper layers when the sparsity is low, and as the sparsity increases, it gradually extends to earlier layers. For example, on the SST-2 task, when the sparsity is set to 20\%, token pruning is not applied to layers 1 to 7. However, when the sparsity increases to 65\%, token pruning is activated in layers 3 to 7, while only layers 1 and 2 are excluded from pruning.

\subsection{Inference latency on Hardware}
\begin{table}[t]
	\begin{center}
	\small
		\begin{tabular}	{@{\hskip0pt}c@{\hskip3pt}c@{\hskip3pt}c@{\hskip3pt}c@{\hskip3pt}c@{\hskip0pt}}
			\hline
			Dataset &\makecell{Input Token\\ length} &BERT &BERT with {\sysname}&\makecell{Latency speedup\\ (Acc. compared to BERT)} \\
			\hline 
			MRPC &128 &84.0 ms&\textbf{29.1 ms}&\textbf{2.9$\times$} (-0.1\%) \\
			RTE & 256&162.8 ms &\textbf{27.8 ms}& \textbf{5.8$\times$} (+2.2\%)\\
			SQuAD & 384& 239.4 ms & \textbf{58.2 ms}&\textbf{4.1$\times$} (-0.2\%) \\
			20News & 512&347.6 ms &\textbf{47.0 ms}&\textbf{7.4$\times$} (+0.3\%)\\
			\hline
		\end{tabular}
	\vspace{2pt}
		\caption{Real inference latency on an 8-core Intel CPU. }
		\label{tbl:latency}
	\end{center}
\end{table}
Finally, we assess the practical efficiency of {\sysname} in resource-limited environments by evaluating BERT on MRPC, RTE, SQuAD, and 20News datasets using an 8-core Intel(R) Xeon(R) CPU@2GHz. The learned ranking masks are applied layer-by-layer to discard tokens, and latency is measured using the high-performance Onnxruntime inference engine~\cite{onnxruntime}. The batch size is set to 1 for simplicity.

As shown in Table~\ref{tbl:latency}, BERT inference without token pruning incurs a high latency on the CPU, particularly for long input sequences. However, {\sysname} significantly reduces this latency by discarding unnecessary tokens. Specifically, {\sysname} delivers inference acceleration of 2.9$\times$, 5.8$\times$, 4.1$\times$, 7.4$\times$ on MRPC, RTE, SQuAD, 20News respectively, with minimal impact on accuracy.
The latency reduction increases with the input length, indicating the big potential of {\sysname} in handling long sequence tasks.

\section{Conclusion}
In this paper, we propose {\sysname}, a novel token pruning approach that leverages $L_0$ regularization to  determine the optimal token removal under a target inference constraint. {\sysname} adopts  hierarchical pruning, where it selects the ideal subset of transformer layers for fine-grained token pruning. 
 Coupled with ranking-aware token importance distillation, {\sysname} significantly enhances the ability of self-attention values to rank token importance, leading to superior accuracy even at high compression ratios. Extensive evaluations on the GLUE benchmark and SQuAD have shown that {\sysname} outperforms existing token pruning and model compression baselines. With the removal of unnecessary tokens, {\sysname} achieves up to 12.6$\times$ FLOP reduction for BERT with less than 1\% drop in accuracy.


{
	\bibliographystyle{ACM-Reference-Format}
	\balance
	\bibliography{ref}
}
\appendix
\newpage
\section{Appendix}

\subsection{Datasets}
Table~\ref{tbl:dataset_summary} summarizes the datasets used in our experiments. Various types of tasks (classification, regression, and QA) with various dataset sizes and input sequence lengths are included to demonstrate the wide applicability of our method.

\begin{table} [h]
	\begin{center}
	\small
		\begin{tabular}	{cccc}
			\hline
			Dataset &Task& \makecell{Avg Sample \\Length}& \makecell{Input Sequence \\Length}\\
			\hline 
			CoLA& Acceptability &11& 64\\
			RTE& NLI & 64&256\\
			QQP& Similarity &30& 128\\
			MRPC& Paraphrase &53& 128\\
			SST-2& Sentiment & 25&64 \\
			MNLI& NLI & 39&128\\
			QNLI& QA& 51&128\\
			STS-B & Similarity&31& 64\\
			SQuAD v2.0 & QA &152& 384\\
			20News& Sentiment&551& 512\\
			\hline
		\end{tabular}
		\caption{ Dataset statistics.  }
		\label{tbl:dataset_summary}
	\end{center}
\end{table}
\subsection{Hyperparameters}

We follow CoFi~\cite{cofi} for setting the total pruning epochs. We use 100 pruning epochs for small GLUE datasets (such as CoLA, RTE, MRPC, and STS-B), and 60 epochs for large  datasets (such as QQP, SST-2, MNLI, QNLI, and 20News). For the warmup stage, we use half of the total  epochs for small datasets, and 25\% for large datasets. 

The hyper-parameter $\lambda$ in Equation~\ref{eq:finalloss} is used to balance the significance of token importance distillation. We use a linear scheduler that reduces the value of $\lambda$ from its initial value throughout the warmup stage. This setting allows the model parameters to adapt more effectively to the teacher's token importance rankings in the early pruning stages, while shifting focus to fine-tuning the model parameters and pruning masks for improved accuracy in the later stages.

We report the hyperparameters used in our experiments in Table ~\ref{tbl:hyperparameter}. $\lambda$ for rank-distillation loss is chosen from \{1e-2, 1e-3, 1e-4\}. LTP, Transkimmer, and CoFiPruning are trained based on their implementation details in the paper and open-source code.

\begin{table} [htb]
	\begin{center}
		\small
		\begin{tabular} {lccc}
		\hline
		Hyperparameter & GLUE & SQuADv2.0 & 20news \\
            \hline
            learning rate & \{1,2,4,8\}e-5 & 4e-5 & 6e-5 \\
            batch size & 32 & 12 & 12 \\
            warmup epoch & 50 (small), 10 (large) & 5 & 10 \\
            total epoch & 100 (small), 60 (large) & 10 & 60 \\
		\hline
		\end{tabular}
		\caption{ Hyperparameters in the experiments.  }
		\label{tbl:hyperparameter}
	\end{center}
\end{table}

\subsection{Case Study}

One example from SQuADv2.0 dataset is presented in Table~\ref{tbl:case_study} to show the reason why we use the attention score from the last layer to distill the attention score of the first few layers. For finetuned $BERT_{base}$ model, the attention score for the answer token ranks 114th in layer 2 among all the attention scores of the total 170 tokens, while in layer 12 it ranks 24th. The answer token is a very important token, and should not be pruned during inference. However, it ranks low in layer 2, so it will be probably pruned for a high sparsity, resulting in a wrong answer. After applying rank distillation, its rank goes higher, making it less likely to be pruned.

\begin{table} [hb]
	\begin{center}
		\small
		\begin{tabular} {cccc}
            \hline
            \multicolumn{4}{p{0.95\columnwidth}}{
                \textbf{Question:} What century did the Normans first gain their separate identity?
            } \\
            \multicolumn{4}{p{0.95\columnwidth}}{
                \textbf{Context:} The Normans (Norman: Nourmands; French: Normands; Latin: Normanni) were the people who in the 10th and 11th centuries gave their name to Normandy, a region in France. They were descended from Norse ("Norman" comes from "Norseman") raiders and pirates from Denmark, Iceland and Norway who, under their leader Rollo, agreed to swear fealty to King Charles III of West Francia. Through generations of assimilation and mixing with the native Frankish and Roman-Gaulish populations, their descendants would gradually merge with the Carolingian-based cultures of West Francia. The distinct cultural and ethnic identity of the Normans emerged initially in the first half of the \textit{10th} century, and it continued to evolve over the succeeding centuries.
            } \\
            \multicolumn{4}{p{0.95\columnwidth}}{
                \textbf{Answer:} 10th.
            } \\
            \hline
            BERT$_{base}$ & \multicolumn{2}{c}{Layer 2} & Layer 12 \\
            Rank of Answer Token & 114 (w/o distill) & 58 (w/ distill) & 24 \\
            \hline
		\end{tabular}
		\caption{ Post-hoc case study of SQuADv2.0 QA task. The answer is highlighted in the context. Rank distillation makes the rank of the answer token higher in the early layer.  }
		\label{tbl:case_study}
	\end{center}
\end{table}

\subsection{Inference on GPU}
 We implement a simple inference version using PyTorch to measure latency on V100 GPU. Our ToP approach demonstrates latency speedup on GPUs, achieving 1.2x speedup on RTE (20.04 ms) and 1.24x speedup on MRPC (11.00 ms), although the acceleration ratio is not as high as on CPUs.

We believe that there is still potential for improving GPU inference acceleration with high-performance inference engines and system optimizations. We found that token pruning requires some memory operations, such as removing tokens with mask value 0 from hidden states. Although this operation requires no computation, it is time-consuming on GPU. In our future work, we plan to utilize high-performance inference engines and leverage system optimizations to achieve greater GPU inference acceleration.

\end{document}